\documentclass[11pt,a4paper]{article}
\usepackage[hyperref]{naaclhlt2018}
\usepackage{times}
\usepackage{latexsym}
\usepackage{graphicx}
\usepackage{amsmath}
\usepackage{multirow}
\usepackage{color}

\newcommand{\comment}[1]{}

\newcommand{\etal}{\mbox{\it et al.}}

\newcommand{\eg}{\mbox{\it e.g.}}

\newcommand{\bi}{\begin{list}{$\bullet$}{
    \setlength{\leftmargin}{1.5 em}
    \setlength{\itemsep}{0 pt}
    \setlength{\topsep}{3 pt}
    \setlength{\parsep}{3 pt}
    \setlength{\partopsep}{0 pt}
    \setlength{\labelwidth}{1 em}
    \setlength{\labelsep}{0.5 em}
    \setlength{\parskip}{0cm}  }}
\newcommand{\ei}{\end{list}}

\newcommand{\BE}{\begin{enumerate}}
\newcommand{\EE}{\end{enumerate}}

\newcommand{\initab}{                           
\begin{tabbing}
XXX \= XXXX \= \kill
}
\newcommand{\begpub}{
\begin{quotation}
\noindent
}

\newcommand{\finpub}{
\end{quotation}
}

\aclfinalcopy 
\def\aclpaperid{1046} 

\title{Semi-Supervised Event Extraction with Paraphrase Clusters}

\author{James Ferguson, Colin Lockard, Daniel S. Weld, and Hannaneh Hajishirzi \\ 
 University of Washington \\
  \texttt{\ \{jfferg, lockardc, weld, hannaneh\}@cs.washington.edu}}

\date{}

\begin{document}

\maketitle

\begin{abstract}
Supervised event extraction systems are limited in their accuracy due to the lack of available training data. We present a method for self-training event extraction systems by bootstrapping additional training data. This is done by taking advantage of the occurrence of multiple mentions of the same event instances across newswire articles from multiple sources. If our system can make a high-confidence extraction of some mentions in such a cluster, it can then acquire diverse training examples by adding the other mentions as well. Our experiments show significant performance improvements on multiple event extractors over ACE 2005 and TAC-KBP 2015 datasets.
\end{abstract}

\section{Introduction}


Event extraction is a challenging task, which aims to discover event triggers in a sentence and classify them by type.  Training an event extraction system requires a large dataset of annotated event triggers and their types in a sentence. Unfortunately, because of the large amount of different event types, each with its own set of annotation rules, such manual annotation is both time-consuming and expensive. As a result, popular event datasets, such as ACE~\cite{ace2005} and TAC-KBP~\cite{Mitamura2015EventNA}, are small (\eg, the median number of positive examples per subtype is only 65 and 86, respectively) and biased
towards frequent event types, such as Attack. 

When an event occurs, there are often multiple parallel descriptions of that event (Figure~\ref{fig:example}) available somewhere on the Web due to the large number of different news sources.
Some  descriptions are simple, explaining in basic language the event that occurred. These are often easier for existing extraction systems to identify.
Meanwhile, other descriptions might use more complex language that falls outside the scope of typical event extractors, but which, if identified, could serve as valuable training data for said systems.

\begin{figure}
\noindent\fbox{%
\parbox{0.95\columnwidth}{%
\small 1) \textbf{LSU} \textit{fires} head coach \textbf{Les Miles} after 12 seasons.\\
2) \textbf{Les Miles} is \textit{out} at \textbf{LSU} after 12 seasons in Baton Rouge.\\
3) On Sunday morning, \textbf{LSU} athletic director Joe Alleva told \textbf{Les Miles} that the coach would \textit{no longer represent} Louisiana State.}
}\vspace{-.2cm}
\caption{Example of a cluster of paraphrases. Shared entities are bolded, and the triggers are italicized. Some, such as the first sentence, are very simple. Others, like the third sentence are more difficult.}
\label{fig:example}
\end{figure}

We automatically generate labeled training data for event trigger identification leveraging this wealth of event descriptions\footnote{The generated data and our code can be found at https://github.com/jferguson144/NewsCluster}.
Specifically, we first group together paraphrases of event mentions. We then use the simple examples in each cluster to assign a label
to the entire cluster. This simplifies the task of extracting events from difficult examples; rather than having to identify whether an event occurs, and which word serves as a trigger for that event, our system needs only to identify the most likely trigger for the given event. Finally, we combine the new examples with the original training set and retrain the event extractor.

Our experiments show that this data can be used with limited amounts of gold data to achieve significant improvement over both standard and neural event extraction systems. In particular, it achieves 1.1 and 1.3 point F1 improvements over a state-of-the-art system in trigger identification on TAC-KBP and ACE data respectively. Moreover, we  show how the benefit of our method varies as a function of the amount of fully-supervised training data and the number of additional heuristically-labeled examples.



\comment{
1) John died yesterday.\\
2) John passed away yesterday.\\
3) John succumbed to cancer yesterday.
}

\vspace{-.1cm}
\section{Approach}
Our goal is to automatically add high quality labeled examples, which can then be used as additional training data to improve the performance of any event extraction model. 
Our data generation process has three steps. The first is to identify clusters of news articles all describing the same event. 
The second step is to run a baseline system over the sentences in these clusters to identify events found in each cluster.
Finally, once we have identified an event in one article in a cluster, our system scans through the other articles in that cluster choosing the most likely trigger in each article for the given event type.

\vspace{-.2cm}
\paragraph{Cluster Articles}
In order to identify groups of articles describing the same event instance, we use an approach inspired by the NewsSpike idea introduced in  \citet{Zhang2015ExploitingPN}. The main intuition is that rare entities that are mentioned a lot on a single date are more indicative that two articles are covering the same event.
%
%
We assign a score, $S$, to each pair of articles, ($a_i$, $a_j$) appearing on the same day, for whether or not they cover the same event, as follows: 
\begin{equation} \label{eq1}
S(a_i, a_j) = \sum\limits_{e \in E_{a_i} \cap E_{a_j}} \frac{\text{count}(e, \text{date}_{a_i,a_j})}{\text{count}(e, \text{corpus}) }, 
\end{equation}
where $E_{a}$ is the list of named entities for the article $a$, and count is the number of times the entity appears on the given date, or in the whole corpus. 
This follows from the intuition above by reducing the weight given to common entities. For example, \textit{United States} appears 367k times in the corpus, so it is not uncommon for it to appear hundreds of times on a single day, and articles mentioning it could be covering completely different topics. Meanwhile \textit{Les Miles} appears only 1.6k times in the corpus, so when there are hundreds of mentions involving \textit{Les Miles} on a single day, it is much more likely that he participated in some event. Accumulating these counts over all shared entities between two articles thus indicates whether the articles are covering the same event.
We then group all articles that cover the same event according to this score into clusters. 


\vspace{-.2cm}
\paragraph{Label Clusters}
Then, given clusters of articles, we run a baseline extractor which was trained on what limited amount of fully-supervised training data is available. The hope is that one or more of a cluster's sentences will use language similar enough to our training data that the extractor can make an accurate prediction. Our system  keeps any cluster in which the baseline system identifies at least some threshold, $\theta_{event}$, of event mentions for a single event type, and labels those clusters with the identified type.

\vspace{-.2cm}
\paragraph{Assign Triggers}
After labeling, the event clusters are comprised of articles in which at least one sentence should contain event mentions of the labeled type. Because most current event extraction systems require labeled event triggers for sentences, we identify those sentences and the event triggers therein so that we can run the baseline systems. 
%
For each sentence we identify the most likely trigger by checking the similarity of the word embeddings to the canonical vector for that event. This vector is computed as the average of the embeddings of the event triggers, $v_t$, in the gold training data:
%
$v_{event} = \frac{1}{|T_{event}|} \sum\limits_{t \in T_{event}}v_t$, where $T_{event}$ is the set of triggers for this event in the gold training data. 
%
If the maximum similarity is greater than some threshold, $\theta_{sim}$, the sentence and the corresponding trigger are added to the training data.


\vspace{-.2cm}
\paragraph{Event Trigger Identification Systems} Event extraction tasks such as ACE and TAC-KBP have frequently been approached with supervised machine learning systems based on hand-crafted features, such as the system adapted from \citet{Li2013JointEE} which we make use of here. Recently, state-of-the-art results have been obtained with neural-network-based systems \cite{Nguyen2016JointEE, Chen2015EventEV, Feng2016ALN}. Here, we make use of two systems whose implementations are publicly available and show that adding additional data would improve their performance. 

The first system is the joint recurrent neural net (JRNN) introduced by \citet{Nguyen2016JointEE}. This model uses a bi-directional GRU layer to encode the input sentence. It then concatenates that with the vectors of words in a window around the current word, and passes the concatenated vectors into a feed-forward network to predict trigger types for each token. Because we are only classifying triggers, and not arguments, we don't include the memory vectors/matrices, which primarily help improve argument prediction, or the argument role prediction steps of that model. 


The second is a conditional random field (CRF) model with the trigger features introduced by \citet{Li2013JointEE}. These include lexical features, such as tokens, part-of-speech tags, and lemmas, syntactic features, such as dependency types and arcs associated with each token, and entity features, including unigrams/bigrams normalized by entity types, and the nearest entity in the sentence. In particular, we use the Evento system from \citet{Ferguson2017TAC}.

\section{Experimental Setup}
\vspace{-.2cm}
\paragraph{Labeled Datasets}
We make use of two labeled datasets: ACE-2005 and TAC-KBP 2015. For the ACE data, we use the same train/development/test split as has been previously used in \cite{Li2013JointEE}, consisting of 529 training documents, 30 development documents, and a test set consisting of 40 newswire articles containing 672 sentences. 
For the TAC-KBP 2015 dataset, we use the official train/test split as previously used in \citet{Peng2016EventDA} consisting of 158 training documents and 202 test documents. 
ACE contains 33 event types, and TAC-KBP contains 38 event types. 
%


For our approach, we use a collection of news articles scraped from the web. These articles were scraped following the approach described in \citet{Zhang2013HarvestingPN}. The process involves collecting article titles from RSS news seeds, and then querying the Bing news search with these titles to collect additional articles. This process was repeated on a daily basis between January 2013 and February 2015, resulting in approximately 70 million sentences from 8 million articles. Although the seed titles were collected during that two year period, the search results include articles from prior years with similar titles, so the articles range from 1970 to 2015.

\vspace{-.2cm}
\paragraph{Evaluation}
We report the micro-averaged F1 scores over all events. A trigger is considered correctly labeled if both its offsets and event type match those of a reference trigger.

\begin{table}[t]
\begin{center}
\resizebox{\columnwidth}{!}{
\begin{tabular}{|c|c|ccc|ccc|}
\hline
\multicolumn{2}{|c|}{}& \multicolumn{3}{|c|}{ACE} & \multicolumn{3}{|c|}{TAC-KBP}\\
\multicolumn{2}{|c|}{}& P & R & F1 & P & R & F1\\
\hline
\multirow{4}{*}{CRF} & 0\% &62.9&70.0&66.3 &53.5&52.3&52.9\\
&10\% &64.5&69.8&67.0&\textbf{59.9}&49.3&\textbf{54.}$\mathbf{1}^*$\\
&20\% &\textbf{65.1}&\textbf{70.2}&\textbf{67.}$\mathbf{6}^*$&59.3&49.2&53.8\\
&30\% &\textbf{65.1}&69.9&67.4&58.1&\textbf{49.4}&53.4\\
\hline
\multirow{4}{*}{JRNN} & 0\% &65.7&72.9&69.1&\textbf{68.8} & 49.2 & 57.3\\
&10\% &67.4&72.7&69.9&65.4 & 52.1 & 58.0\\
&20\% &\textbf{67.6}&\textbf{73.5}&\textbf{70.}$\mathbf{4}^*$&65.3 & 52.8& \textbf{58.}$\mathbf{4}^*$\\
&30\% &67.5&73.3&70.3&64.7&\textbf{52.9}&58.2\\
\hline \hline
\multicolumn{2}{|c|}{HNN} & 84.6 & 64.9 & 73.4 & - & - & -\\
\hline
\multicolumn{2}{|c|}{SSED} &-&-&-& 69.9 &48.8 & 57.5\\
\hline

\end{tabular}
}
\end{center}\vspace{-.2cm}
\caption{Results after adding varying amounts of automatically-generated news data. Percentages indicate the amount of additional data relative to the size of the gold training data. Using a modest amount of semi-supervised data improves extractor performance on both ACE \&\ TAC-KBP events. * indicates that the difference in F1 relative to training with just the gold data is statistically significant $(p < 0.05)$.}
\label{fig:scaleResults}
\end{table}

\vspace{-.2cm}
\paragraph{Implementation details}
For creating the automatically-generated data, we set thresholds  $\theta_{event}$ and $\theta_{sim}$ to 2 and 0.4 respectively, which were selected according to validation data. We use CoreNLP \cite{corenlp} for named entity recognition, and we use a pre-trained Word2Vec model \cite{Mikolov2013DistributedRO} for the vector representations. 

For the JRNN model, we follow the parameter settings of \cite{Nguyen2016JointEE} and use a context window of 2 for context words, and a feed-forward neural network with one hidden layer for trigger prediction with hidden layer size of 300. Finally, for training, We apply the stochastic gradient descent algorithm with mini-batches of size 50 and the AdaDelta update rule \cite{zeiler} with L$_2$ regularization. 
For the CRF model, we maximize the conditional log likelihood of the training data with a loss function via softmax-margin \cite{Gimpel2010}. We optimize using AdaGrad \cite{Duchi2011} with L$_2$ regularization.

\section{Experiments}

\vspace{-.2cm}
\paragraph{Varying Amounts of Additional Data}
In this section we show that the addition of automatically-generated training examples improves
the performance of both systems we tested it on. We sample examples from the 
automatically-generated data, limiting the total number of positive examples to a specific number. In order to avoid biasing the system in favor of a
specific event type, we ensure that the additional data has a uniform distribution of event types. We run 10 trials at each point, and report
 average results.

Table~\ref{fig:scaleResults} reports the results of adding varying amounts of our generated data to both CRF and JRNN systems. We observe that  that adding any amount of heuristically-generated data improves performance. Optimal performance, however, is achieved fairly early in both datasets. This is likely due to the domain mismatch between the gold and additional data. 
For reference purposes, we also include the result of using the HNN model from \cite{Feng2016ALN} and the SSED system from \cite{Sammons2015}, which are the best reported results on the ACE-2005 and TAC-KBP 2015 corpora respectively. These systems could also benefit from our additional data since our approach is system independent. 

 \begin{figure}[t]
 \centering
\includegraphics[scale=0.5]{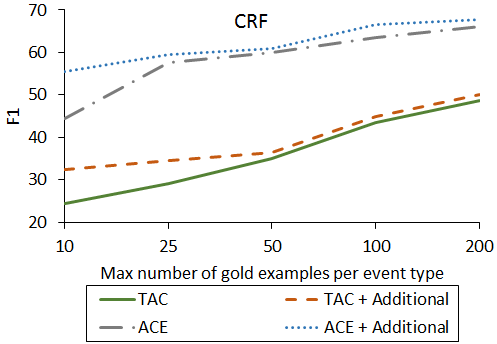}
\vspace*{-0.2cm}
\caption{Adding a reasonable amount (200 examples per event) of semi-supervised data on top of limited amounts of gold training data improves performance across the board, but the gain is dramatic when the number of supervised examples is extremely small.}
\label{fig:limitResults}
\end{figure}
 
 \vspace{-.2cm}
 \paragraph{Varying Amounts of Supervised Data}
In this section we evaluate how the benefit of adding semi-supervised data varies given different amounts of  gold (supervised) data to start. We conjecture that semi-supervision will be more beneficial when gold data is very limited, but the conclusion isn't obvious, since semi-supervision is more likely to add noisy examples in this case.  
Specifically, we limit the number of positive gold examples for each event by randomly sampling the overall set. We then add in the same amount of automatically-generated data
to each trial. We again run 10 trials for each size, and report the average.

The results for this experiment using the CRF model can be seen in figure \ref{fig:limitResults}: training with large amounts of semi-supervised data improves performance considerably when limited gold training data is available, but those gains diminish with more high-quality supervised data. We observe the same trend for the JRNN system as well.

\vspace{-.2cm}
\paragraph{Discussion}

We randomly selected 100 examples from the automatically-generated data and manually annotated them. For each example that did not contain a correctly labeled event mention, we further annotated where in the pipeline an error occurred to cause the incorrect labeling. This breakdown can be seen in table \ref{fig:ErrorAnalysis}. As observed in the table, the errors are mainly due to the incorrect event identification or trigger assignment.  

Incorrect clustering refers to cases in which a sentence does not cover the same topic as other sentences in its cluster. This was primarily caused by entities participating in multiple events around the same time period. For example, this occurred in sentences from the 2012 US presidential election coverage involving Barack Obama and Mitt Romney. 

\begin{table}[t]
\begin{tabular}{|c|l|c|}
\hline
\multicolumn{2}{|c|}{Correct} & 72 \\
\hline
\multirow{3}{*}{Incorrect }& clustering & 5 \\
&event identification & 13\\
&trigger assignment & 10\\
\hline
\end{tabular}
\caption{The results of manually labeling 100 examples that were automatically-generated using JRNN as the supervised system.}
\label{fig:ErrorAnalysis}
\end{table}

Incorrect event identification refers to clusters that were incorrectly labeled by the supervised system. The primary reason for these errors is due to domain mismatch between the news articles and the gold data. For example, our system identifies the token \textit{shot} in \textit{Bubba Watson shot a 67 on Friday} as an attack event trigger. Because the gold data does not contain examples involving sports, the baseline system mistakenly identifies a paraphrase of the above sentence as an attack event, and our system is not able to fix that mistake. However, this problem can be solved by training the baseline extractor on the same domain as the additional data.

Incorrect trigger assignment refers to errors in which a sentence is correctly identified as containing an event mention, but the wrong token is selected as a trigger. The most common source of this error is tokens that are strongly associated with multiple events. For example, \textit{shooting} is strongly associated with both attack and die events, but only actually indicates an attack event.

Looking through the correct examples, the data collection process is able to identify uncommon triggers that do not show up in the baseline training data. For example, it correctly identifies  ``offload'' as a trigger for Transfer-Ownership in \textit{Barclays is to offload part of its Spanish business to Caixabank}. Despite the trigger identification step having no context awareness, the process is also able to correctly identify triggers that rely on context, such as ``contributions'' triggering Transfer-Money in \textit{Chatwal made \$188,000 of illegal campaign contributions to three U.S. candidates via straw donors}.

\section{Related Work}

A challenge in event extraction is the relatively small number of labeled training examples available. Researchers have dealt with this by framing event extraction in a way that allows them to rely heavily on systems built for dependency parsing \cite{McClosky2011EventEA} and semantic role labeling \cite{Peng2016EventDA}. Unlike these researchers, we join a line of work that attempts to directly harvest additional training examples for use in traditional event extraction systems.

\comment{
Others have explored methods of trigger identification via similarity to a small number of seed triggers from event definitions\cite{Bronstein2015SeedBasedET}.
}

Distant supervision is one source of additional data that has been successfully applied to relation extraction tasks~\cite{Riedel2010ModelingRA,Hoffmann2011KnowledgeBasedWS, Mintz2009DistantSF}, which align a background knowledge base to an accompanying corpus of natural language documents. For event extraction, such data sources are not as easily available since there are no pre-existing stores of tuples of attacks, movements or meetings. 

Other work has generated  additional data by using a pattern-based model of event mentions and bootstrapping on top of a small set of seed examples. \citet{Huang2012Bootstrap} begin with a set of nouns that are specific to certain event roles and extract patterns based on the contexts in which those words appear. \citet{Li2014SemiChinese} extracted additional patterns using various event inference mechanisms.

The work most similar to ours is that of \citet{Liao2010FilteredRF,Liao2011CanDS} to identify articles from a  corpus which described the same event instances found in training examples. 
These articles are then used in self-training an ACE-trained system after being filtered to select passages with consistent roles and triggers. Their method provides a 2.7 point boost to F1, but their baseline system results are much lower than ours (54.1 vs 69.1) and it is unclear what improvement their method would have on a state-of-the-art extractor. In addition, their system attempts to identify relevant articles that describe event instances already present in their training data, while we attempt to find clusters of sentences describing a common event, at least one of which we can confidently label.

The use of parallel news streams to acquire event extraction training data in an unsupervised fashion was explored in \cite{Zhang2015ExploitingPN}, whose clustering methods we have adapted here. Unlike Zhang \etal, we have a defined event ontology for which we are acquiring data, rather than attempting to learn event types from the data. Furthermore, we use an extractor trained on fully-supervised examples to filter clusters, in contrast to Zhang \etal, whose method is completely unsupervised, which allows us to relax some of the assumptions made by Zhang \etal\  and consider ``spikes'' of individual entities as opposed to pairs.

\vspace{-.1cm}
\section{Conclusion}
We present a method for self-training event extraction systems by taking advantage of parallel mentions of the same event instance in newswire text. By examining clusters of sentences which produce at least two extractions of the same event type and assigning a trigger label to each sentence via word embedding similarity, we add diverse training examples to our dataset. Our experiments show a 1.3 point F1 increase in trigger labeling for a state-of-the-art baseline system on ACE, and a 1.1 point increase on TAC-KBP. For future research, this work can be applied to arbitrary event extraction models to improve performance, or make up for a lack of training data. The code and data are publicly available at our github repository.

\section*{Acknowledgements}
This research was supported in part by ONR grants N00014-11-1-0294 and N00014-15-1-2774, DARPA contract FA8750-13-2- 0019, NSF (IIS-1616112,  IIS-1420667, IIS-1703166), ARO grant W911NF13-1-0246 and the WRF/T.J. Cable Professorship. We would like to thank Stephen Soderland for many helpful discussions and our anonymous reviewers for their comments.

\bibliography{naaclhlt2018}

\begin{thebibliography}{24}
\expandafter\ifx\csname natexlab\endcsname\relax\def\natexlab#1{#1}\fi

\bibitem[{Chen et~al.(2015)Chen, Xu, Liu, Zeng, and Zhao}]{Chen2015EventEV}
Yubo Chen, Liheng Xu, Kang Liu, Daojian Zeng, and Jun Zhao. 2015.
\newblock Event extraction via dynamic multi-pooling convolutional neural
  networks.
\newblock In \emph{ACL}.

\bibitem[{Duchi et~al.(2011)Duchi, Hazan, and Singer}]{Duchi2011}
John Duchi, Elad Hazan, and Yoram Singer. 2011.
\newblock Adaptive subgradient methods for online learning and stochastic
  optimization.
\newblock In \emph{Journal of Machine Learning Research}.

\bibitem[{Feng et~al.(2016)Feng, Huang, Tang, Ji, Qin, and Liu}]{Feng2016ALN}
Xiaocheng Feng, Lifu Huang, Duyu Tang, Heng Ji, Bing Qin, and Ting Liu. 2016.
\newblock A language-independent neural network for event detection.
\newblock In \emph{ACL}.

\bibitem[{Ferguson et~al.(2017)Ferguson, Lockard, Hawkins, Soderland,
  Hajishirzi, and Weld}]{Ferguson2017TAC}
James Ferguson, Colin Lockard, Natalie Hawkins, Stephen Soderland, Hannaneh
  Hajishirzi, and Daniel~S. Weld. 2017.
\newblock University of washington tac-kbp 2016 system description.
\newblock In \emph{TAC-KBP}.

\bibitem[{Gimpel and Smith(2010)}]{Gimpel2010}
Kevin Gimpel and Noah~A. Smith. 2010.
\newblock Softmaxmargin crfs: Training log-linear models with cost functions.
\newblock In \emph{HLT-NAACL}.

\bibitem[{Hoffmann et~al.(2011)Hoffmann, Zhang, Ling, Zettlemoyer, and
  Weld}]{Hoffmann2011KnowledgeBasedWS}
Raphael Hoffmann, Congle Zhang, Xiao Ling, Luke~S. Zettlemoyer, and Daniel~S.
  Weld. 2011.
\newblock Knowledge-based weak supervision for information extraction of
  overlapping relations.
\newblock In \emph{ACL}.

\bibitem[{Huang and Riloff(2012)}]{Huang2012Bootstrap}
Ruihong Huang and Ellen Riloff. 2012.
\newblock Bootstrapped training of event extraction classifiers.
\newblock In \emph{EACL}.

\bibitem[{Li et~al.(2014)Li, Zhu, and Zhou}]{Li2014SemiChinese}
Peifeng Li, Qiaoming Zhu, and Guodong Zhou. 2014.
\newblock Employing event inference to improve semi-supervised chinese event
  extraction.
\newblock In \emph{COLING}.

\bibitem[{Li et~al.(2013)Li, Ji, and Huang}]{Li2013JointEE}
Qi~Li, Heng Ji, and Liang Huang. 2013.
\newblock Joint event extraction via structured prediction with global
  features.
\newblock In \emph{ACL}.

\bibitem[{Liao and Grishman(2010)}]{Liao2010FilteredRF}
Shasha Liao and Ralph Grishman. 2010.
\newblock Filtered ranking for bootstrapping in event extraction.
\newblock In \emph{COLING}.

\bibitem[{Liao and Grishman(2011)}]{Liao2011CanDS}
Shasha Liao and Ralph Grishman. 2011.
\newblock Can document selection help semi-supervised learning? a case study on
  event extraction.
\newblock In \emph{ACL}.

\bibitem[{Manning et~al.(2014)Manning, Surdeanu, Bauer, Finkel, Bethard, and
  McClosky}]{corenlp}
Christopher~D. Manning, Mihai Surdeanu, John Bauer, Jenny Finkel, Steven~J.
  Bethard, and David McClosky. 2014.
\newblock The {Stanford} {CoreNLP} natural language processing toolkit.
\newblock In \emph{Association for Computational Linguistics (ACL) System
  Demonstrations}, pages 55--60.

\bibitem[{McClosky et~al.(2011)McClosky, Surdeanu, and
  Manning}]{McClosky2011EventEA}
David McClosky, Mihai Surdeanu, and Christopher~D. Manning. 2011.
\newblock Event extraction as dependency parsing.
\newblock In \emph{ACL}.

\bibitem[{Mikolov et~al.(2013)Mikolov, Sutskever, Chen, Corrado, and
  Dean}]{Mikolov2013DistributedRO}
Tomas Mikolov, Ilya Sutskever, Kai Chen, Gregory~S. Corrado, and Jeffrey Dean.
  2013.
\newblock Distributed representations of words and phrases and their
  compositionality.
\newblock \emph{CoRR}, abs/1310.4546.

\bibitem[{Mintz et~al.(2009)Mintz, Bills, Snow, and
  Jurafsky}]{Mintz2009DistantSF}
Mike Mintz, Steven Bills, Rion Snow, and Daniel Jurafsky. 2009.
\newblock Distant supervision for relation extraction without labeled data.
\newblock In \emph{ACL/IJCNLP}.

\bibitem[{Mitamura et~al.(2015)Mitamura, Yamakawa, Holm, Song, Bies, Kulick,
  and Strassel}]{Mitamura2015EventNA}
Teruko Mitamura, Yukari Yamakawa, Susan Holm, Zhiyi Song, Ann Bies, Seth
  Kulick, and Stephanie Strassel. 2015.
\newblock Event nugget annotation: Processes and issues.

\bibitem[{Nguyen et~al.(2016)Nguyen, Cho, and Grishman}]{Nguyen2016JointEE}
Thien~Huu Nguyen, Kyunghyun Cho, and Ralph Grishman. 2016.
\newblock Joint event extraction via recurrent neural networks.
\newblock In \emph{HLT-NAACL}.

\bibitem[{Peng et~al.(2016)Peng, Song, and Roth}]{Peng2016EventDA}
Haoruo Peng, Yangqiu Song, and Dan Roth. 2016.
\newblock Event detection and co-reference with minimal supervision.
\newblock In \emph{EMNLP}.

\bibitem[{Riedel et~al.(2010)Riedel, Yao, and McCallum}]{Riedel2010ModelingRA}
Sebastian Riedel, Limin Yao, and Andrew McCallum. 2010.
\newblock Modeling relations and their mentions without labeled text.
\newblock In \emph{ECML/PKDD}.

\bibitem[{Sammons et~al.(2015)Sammons, Peng, Song, Upadhyay, Tsai, Reddy, Roy,
  and Roth}]{Sammons2015}
Mark Sammons, Haoruo Peng, Yangqiu Song, Shyam Upadhyay, Chen-Tse Tsai,
  Pavankumar Reddy, Subhro Roy, and Dan Roth. 2015.
\newblock Illinois ccg tac 2015 event nugget, entity discovery and linking, and
  slot filler validation systems.
\newblock In \emph{TAC}.

\bibitem[{Walker et~al.(2006)Walker, Strassel, Medero, and Maeda}]{ace2005}
Christopher Walker, Stephanie Strassel, Julie Medero, and Kazuaki Maeda. 2006.
\newblock \emph{ACE 2005 Multilingual Training Corpus LDC2006T06}.
\newblock Linguistic Data Consortium, Philadelphia.

\bibitem[{Zeiler(2012)}]{zeiler}
Matthew~D. Zeiler. 2012.
\newblock {ADADELTA:} an adaptive learning rate method.
\newblock \emph{CoRR}, abs/1212.5701.

\bibitem[{Zhang et~al.(2015)Zhang, Soderland, and Weld}]{Zhang2015ExploitingPN}
Congle Zhang, Stephen Soderland, and Daniel~S. Weld. 2015.
\newblock Exploiting parallel news streams for unsupervised event extraction.
\newblock \emph{TACL}, 3:117--129.

\bibitem[{Zhang and Weld(2013)}]{Zhang2013HarvestingPN}
Congle Zhang and Daniel~S. Weld. 2013.
\newblock Harvesting parallel news streams to generate paraphrases of event
  relations.
\newblock In \emph{EMNLP}.

\end{thebibliography}
\bibliographystyle{acl_natbib}

\end{document}